\documentclass[11pt,a4paper]{article}
\usepackage[hyperref]{acl2019}
\usepackage{times}
\usepackage{latexsym}

\usepackage{url}

\usepackage{adjustbox}
\usepackage{amsmath}
\usepackage{booktabs}
\usepackage{comment}
\usepackage{dirtytalk}
\usepackage{enumitem}
\usepackage{microtype}
\usepackage{siunitx}
\newcommand{\term}[1]{\textbf{#1}}
\usepackage{longtable}
\usepackage{supertabular}
\usepackage{tipa}
\usepackage{mathptmx}
\usepackage{multirow}

\DeclareMathAlphabet{\mathcal}{OMS}{cmsy}{m}{n}
\DeclareMathAlphabet{\mathbb}{U}{msb}{m}{n}

\usepackage{cleveref}

\newcommand{\prefix}[1]{/\textipa{#1}/-}
\newcommand{\suffix}[1]{-/\textipa{#1}/}

\DeclareMathOperator{\Entr}{H}
\DeclareMathOperator{\MI}{I}
\DeclareMathOperator{\Unc}{U}
\DeclareMathOperator{\softmax}{softmax}
\DeclareMathOperator{\LSTM}{LSTM}

\usepackage[disable]{todonotes}
\makeatletter
\newcommand*\iftodonotes{\if@todonotes@disabled\expandafter\@secondoftwo\else\expandafter\@firstoftwo\fi}  
\makeatother

\newcommand{\note}[4][]{\todo[author=#2,color=#3,size=\scriptsize,fancyline,caption={},#1]{#4}} 

\newcommand{\arya}[2][]{\note[#1]{arya}{cyan!40}{#2}}

\newcommand{\tiago}[2][]{\note[#1]{tiago}{green!40}{#2}}

\newcommand{\damian}[2][]{\note[#1]{damian}{yellow!40}{#2}}

\newcommand{\brian}[2][]{\note[#1]{brian}{violet!40}{#2}}

\def\Snospace~{\S{}}

\newcommand{\word}[1]{\textit{#1}}

\aclfinalcopy 
\def\aclpaperid{525} 

\title{Meaning to Form: 
Measuring Systematicity as Information}

\author{Tiago Pimentel$^\clubsuit$ 
~\; Arya D. McCarthy$^\heartsuit$ ~\; Dami\'{a}n E. Blasi$^\spadesuit$ ~\; Brian Roark$^\diamondsuit$ ~\; Ryan Cotterell$^{\heartsuit\dagger}$\\
$^\clubsuit$Kunumi,~\;~$^\heartsuit$Johns Hopkins University,~\;~%
$^\spadesuit$University of Z\"{u}rich \& MPI SHH,~\;~\\ $^\diamondsuit$Google,~\;~%
$^\dagger$University of Cambridge\\
{\tt tiago.pimentel@kunumi.com},~\;~ {\tt arya@jhu.edu},~\;~ 
{\tt  damian.blasi@uzh.ch},\\
{\tt roarkbr@gmail.com},~\;~ {\tt rdc42@cam.ac.uk}
}  

\date{}

\begin{document}
\maketitle 
\begin{abstract}
\brian[disable]{simultaneously philosophically opposed to whiff's of Harry Potter in my work (see title) and philosophically opposed to any form of philosophical opposition, so... just sayin'.  won't make a stink but would be fine with a less beWITCHing name...}
  A longstanding debate in semiotics centers on the relationship between linguistic signs and their corresponding semantics: is there an arbitrary relationship between a word form and its meaning, or does some systematic phenomenon pervade? For instance,
  does the character bigram \word{gl} have any systematic relationship to the meaning of words like \word{glisten}, \word{gleam} and \word{glow}?
  In this work, we offer a holistic quantification of the systematicity of the sign using mutual information and recurrent neural networks.
  We employ these in a data-driven and massively multilingual approach to the question, examining 106 languages. 
  We find a statistically significant reduction in entropy when modeling a word form conditioned on its semantic representation.
  Encouragingly, we also recover well-attested English examples of systematic affixes.
  We conclude with the meta-point: Our approximate effect size \arya[disable]{did we compute this?}  (measured in bits) is quite small---
  despite some amount of systematicity between form and meaning, an arbitrary relationship and its resulting benefits dominate human language.
\end{abstract}

\section{Introduction}

\begin{figure}
  \begin{center}
    \includegraphics[width=\columnwidth]{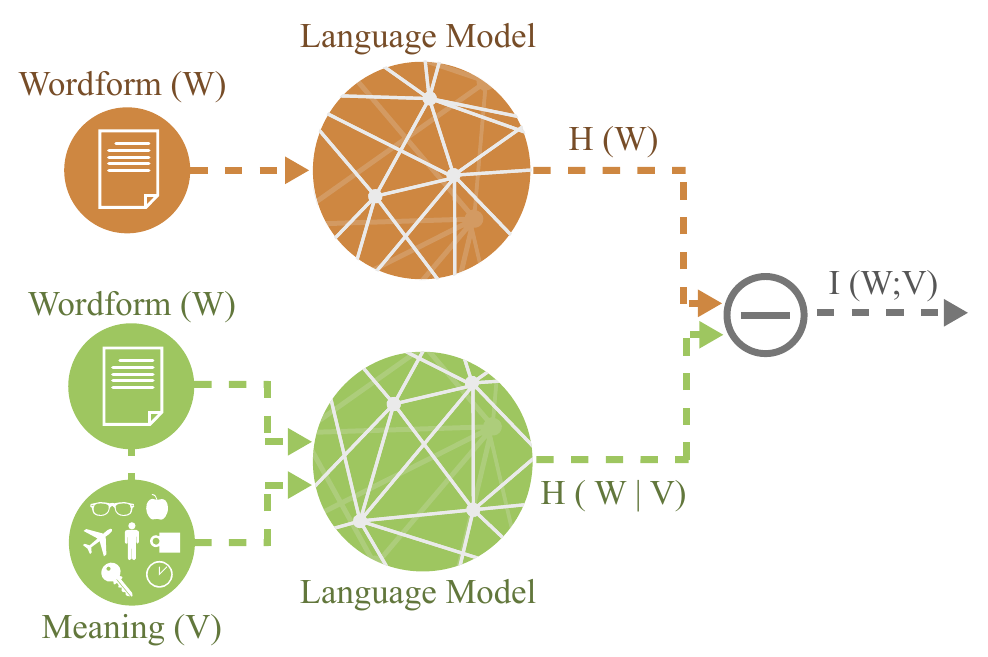}
  \end{center}
  \caption{We use two independent language models to estimate the mutual information between word forms and meaning---i.e. systematicity, as per our definition. The language models provide upper bounds on $\Entr(W)$ and $\Entr(W \mid V)$, which can be used to  estimate  $\MI(W; V)$. This estimate is as good as the upper bounds are tight---see discussion in \autoref{sec:choosing_q}.} \label{fig:schematic}
\end{figure}

\citet{saussure1916course} expounded on the \term{arbitrariness of the sign}. Seen as a critical facet of human language \citep{hockett1960origin}, the idea posits that a sign in human language (a word, in our inquiry) is structured at two levels: the \term{signified}, which captures its meaning, and the \term{signifier}, which has no meaning but manifests the form of the sign. Saussure himself, however, also documented instances of sound symbolism in language \citep{saussure1912}. In this paper, we present computational evidence of relevance to both aspects of Saussure's work.
 
 While dominant among linguists, arbitrariness has been subject to both long theoretical debate \cite{wilkins1974essay, eco1995search, johnson2004systematicity, pullum2007systematicity} and numerous empirical and experimental studies \citep{hutchins1998psychological, bergen2004psychological, monaghan2011arbitrariness, abramova2016questioning, blasi2016sound, gutierrez2016finding, dautriche2017wordform}. Taken as a whole, these studies suggest non-trivial interactions in the form--meaning interface between 
the signified and the signifier \cite{dingemanse2015iconicity}.

Although the new wave of studies on form--meaning associations range across multiple languages, methods and working hypotheses, they all converge on two important dimensions:
\begin{enumerate}[nosep]
    \item The description of meaning is parameterized with pre-defined labels---e.g., by using existing ontologies like \citet{list2016concepticon}.
    \item The description of forms is restricted to the presence, absence or sheer number of occurrence of particular units (such as phones, syllables or handshapes).
\end{enumerate}
%
We take an information-theoretic approach to quantifying the relationship between form and meaning using flexible representations in both domains, rephrasing the question of systematicity: 
\emph{How much does certainty of one reduce uncertainty of the other?}
This gives an operationalization as the \term{mutual information} between form and meaning, when treating both as random variables---the signifier as a word's phone string representation in the International Phonetic Alphabet (IPA), and the signified as a distributed representation \citep{mikolov2013efficient} for that word's lexical
semantics, devoid of morphological or other subword information. We show how to estimate mutual information as the difference in entropy of two phone\damian[disable]{sorry for the quibble, but wouldn't you call this phone or segment rather than phoneme? we are talking about IPA symbols, not language-specific phonemes, no?}\arya[disable]{@Tiago---how strict is the IPA for these datasets?}\tiago[disable]{I think we should change to phones as Damian said. Ryan's note on another paper: We say that $\Sigma$ is the set of \emph{phones}, rather than the set of \emph{phonemes}, as a phonotactic model must account for allophony.}-level LSTM language models---one of which is conditioned on the semantic representation. This operationalization, depicted in \autoref{fig:schematic}, allows us to express the {\em global} effect of meaning on form in vocabulary datasets with wide semantic coverage. 


In addition to this lexicon-level characterization of systematicity, 
we also show that this paradigm can be leveraged for studying more narrowly-defined form-meaning associations such as \term{phonesthemes}---submorphemic, meaning-bearing units---in the style of \citet{gutierrez2016finding}. 
These short sound sequences typically suggest some aspect of meaning in the words that contain them, like \emph{-ump} for rounded things in English.
Previous computational studies, whether focusing on characterizing the degree of systematicity  \citep{monaghan2014arbitrary, monaghan2014systematicity, monaghan2011arbitrariness, shillcock2001filled}, discovering phonesthemes \citep{liu2018discovering}, or both \citep{gutierrez2016finding}, have invariably framed systematicity in terms of distances and/or similarities--the relation between word-form distance/similarity on the one hand (e.g., based on string edit distance) and semantic distance/similarity on the other (e.g., as defined within a semantic vector space).  Our methods have the virtue of not relying on some pre-defined notion of similarity or distance in either domain for our measurement of systematicity.


Empirically, we focus on two experimental regimes. 
First, we focus on a large \tiago[disable]{Is CELEX concept aligned? I don't think so.}
corpus (CELEX) of phone transcriptions in Dutch, English, and German.
In these three languages, we find a significant yet small mutual information even when controlling for grammatical category. 
Second, we perform a massively multilingual exploration of sound--meaning systematicity (\autoref{sec:identifying-systematicity}) on the NorthEuraLex corpus \cite{dellert2017northeuralex}.
This corpus contains expanded Swadesh lists in 106 languages using a unified alphabet of phones. It contains 1016 words in each language, which is often not enough to detect systematicity---we trade the coverage of CELEX for the breadth of languages. Nevertheless, using our information-theoretic operationalization, in most of the languages considered (87 of 106), we find a statistically significant reduction in entropy of phone \tiago[disable]{Should this change to phone as well?} language modeling by conditioning on a word's meaning (\autoref{sec:northeuralex})\damian{@tiago: maybe give a range of mean U for these languages?}\tiago{We added the effect sizes there}. \arya{Sure, but also add it here.} 
Finally, we find a weak positive correlation between our computed mutual information and human judgments of form--meaning relatedness.

\section{Systematic form-meaning associations}
\subsection{Arbitrariness}
The lack of a forceful association between form and meaning is regarded as a design feature of language \cite{hockett1960origin}. This arbitrariness of the sign is thought to provide a flexible and efficient way for encoding new referents \citep{monaghan2011arbitrariness}. It has been claimed that it enhances learnability because newly acquired concepts can be paired to any word, instead of devising the word that properly places the concept in one's constellation of concepts \citep{gasser2005iconicity}, and that it facilitates mental processing compared to an icon-based symbol system, in that the word--meaning map can be direct \citep{lupyan2011evocative}. Most importantly, decoupling form from meaning allows communication about things that are not directly grounded in percepts \citep{clark1998magic, dingemanse2015iconicity}. This opens the door to another of \citet{hockett1960origin}'s design features of language: duality of patterning \citep{martinet1949double}, the idea that language exists on the level of meaningless units (the \emph{distinctive}; typically phonemes) composed to form the level of meaningful units (the \emph{significant}; typically morphemes).
\subsection{Non-arbitrariness and systematicty}\label{sec:sys-icon-arb}

Contemporary research has established that non-arbitrary form-meaning associations in vocabulary are more common and diverse than previously thought \cite{dingemanse2015iconicity}. Some non-arbitrary associations might be found repeatedly across unrelated languages presumably due to species-wide cognitive biases \cite{blasi2016sound}, others are restricted to language-specific word classes that allow for more or less transparent iconic mappings -- so-called {\em ideophones}, see Dingemanse \shortcite{dingemanse2012advances,dingemanse2018redrawing} -- and yet others might emerge from properties of discourse and usage rather than meaning per se \cite{piantadosi2011word}.

{\bf Systematicity} is meant to cover all cases of non-arbitrary form-meaning associations of moderate to large presence in a vocabulary within a language \cite{dingemanse2015iconicity}. In morphology-rich languages, systematic patterns are readily apparent: for instance, across a large number of languages recurring TAM markers or transitivity morphemes could be used to detect verbs, whereas case markers or nominalizing morphemes can serve as a cue for nouns. Yet a sizable portion of research on systematicity is geared towards subtle patterns at the word root level, beyond any ostensive rules of grammar.

By and large, systematicity is hailed as a trait easing language acquisition. It reduces the radical uncertainty humans find when first encountering a new word by providing clues about category and meaning \citep{monaghan2014systematicity}. Systematic patterns can display a large scope within a language: for instance, systematic associations distinguishing nouns from verbs have been found in every language where a comparison was performed systematically  \citep[e.g.][]{monaghan2007phonological}. 
But at its extreme, systematicity would manifest as an ontology encoded phonetically, e.g., all plants begin with the letter `g', and animals with the letter `z' \citep{wilkins1974essay, eco1995search}. As \citet{dingemanse2015iconicity} note, a system of similar forms expressing similar meanings \say{would lead to high confusability of the very items most in need of differentiation}.




\subsection{Phonesthemes}

One particular systematic pattern comes in the form of \term{phonesthemes} \citep{firth1964tongues}. These are  submorphemic and mostly unproductive affixal units, usually flagging a relatively small semantic domain. A classic example in English is \emph{gl-}, a prefix for words relating to light or vision, e.g. \emph{glimmer}, \emph{glisten}, \emph{glitter}, \emph{gleam}, \emph{glow} and \emph{glint} \citep{bergen2004psychological}. 



Phonesthemes have psychological import; they can be shown to accelerate reaction times in language processing \citep{hutchins1998psychological, bergen2004psychological, magnus2000s}. 
They have been attested in English \citep{wallis1699, firth1930speech, marchand1959phonetic, bolinger1949sign, bolinger2014language}, Swedish \citep{abelin1999studies}, Japanese \citep{hamano1998sound}, Ojibwa  \citep{rhodes1981semantics}, Hmong \citep{ratliff1992meaningful}, and myriad Austronesian languages \citep{mccune1985internal, blust1988austronesian}. 
In fact, as \citet{bergen2004psychological} notes, \say{every systematic study of a particular language has produced results suggesting that that language has phonesthemes}. \citet{liu2018discovering} survey computational approaches for identifying phonesthemes.

\section{Estimating Systematicity with Information Theory} \label{sec:methodology}

\subsection{Notation and formalization}

 Following \citet{shillcock2001filled}, we define a sign as a tuple \((\mathbf{v}^{(i)}, \mathbf{w}^{(i)})\) of a word's distributional semantic representation (a vector) and its \damian[disable]{see my comment above: is it really phonological?}\tiago[disable]{If someone knows the equivalent of phonological please fill it here. phoneme $\rightarrow$ phone ; phonological $\rightarrow$ ???} phone string representation (a word form). For a natural language with a set of phones \(\Sigma\)  (including a special end-of-string token), we take the space of word forms to be \(\Sigma^*\), with \(\mathbf{w}^{(i)} \in \Sigma^*\). We treat the semantic space  as a high-dimensional real vector space \(\mathbb{R}^d\), with \(\mathbf{v}^{(i)} \in \mathbb{R}^d\). The particular \(\mathbf{v}^{(i)}\) and \(\mathbf{w}^{(i)}\) are instances of random variables \(V\) and \(W\).
 
 Further, we want to hunt down potential phonesthemes; we define these to be phone sequences which, compared to others of their length, have a larger mutual information with their meaning. We eliminate positional confounds by examining only words' prefixes \(\mathbf{w}_{<k}\) and suffixes \(\mathbf{w}_{>k}\).\footnote{ In line with, e.g., \citet{cucerzan2003minimally}, we treat affixes as word-initial or word-final sequences, regardless of their status as attested morphological entities.}

\subsection{A variational upper bound}

Entropy, the workhorse of information theory, captures the uncertainty of a probability distribution. In our language modeling case, the quantity is
\begin{equation}
    \Entr(W) \equiv \sum_{\mathbf{w} \in \Sigma^*} \Pr(\mathbf{w})  \log \frac{1}{\Pr(\mathbf{w})}\text{.} \label{eq:entropy}
\end{equation}
Entropy is the average number of bits required to represent a string in the distribution, under an optimal coding scheme.
When computing it, we are faced with two problems: We do not know the distribution over word-forms $\Pr(W)$ and, even if we did, computing \autoref{eq:entropy} requires summing over the infinite set of possible strings $\Sigma^*$.

We follow \newcite{brown1992estimate} in tackling these problems together.
Approximating $\Pr(W)$ with any known distribution $Q(W)$, we get a variational upper bound on $\Entr(W)$ from their cross-entropy, i.e.
\begin{subequations}
  \begin{align}
\Entr(W) &\leq \Entr_Q(W) \label{eq:inequality} \\
	&= \sum_{\mathbf{w} \in \Sigma^*} \Pr(\mathbf{w}) \log \frac{1}{Q(\mathbf{w})}\text{.} \label{eq:cross-entropy}
  \end{align}
\end{subequations}
\autoref{eq:cross-entropy} still requires knowledge of $\Pr(W)$ and involves an infinite sum, though. Nonetheless, we can use a finite set $\tilde{\mathbf{W}}$ of samples from $\Pr(W)$ to get an empirical estimate of this value.
\begin{equation}\label{eq:empirical}
  \Entr_Q(W) \approx \frac{1}{N} \sum_{i=1}^N \log \frac{1}{Q\left(\tilde{\mathbf{w}}^{(i)}\right)}, \quad \tilde{\mathbf{w}}^{(i)} \in \tilde{\mathbf{W}} \sim \Pr(W)
\end{equation}
with equality if we let $N \rightarrow \infty$.\footnote{ This is a direct consequence of the law of large numbers.} We now use \autoref{eq:empirical} as an estimate for the entropy of a lexicon.

\paragraph{Conditional entropy}
Conditional entropy reflects the average \emph{additional} number of bits needed to represent an event, given knowledge of another random variable.
If \(V\) completely determines \(W\), then the quantity is \(0\). Conversely, if the variables are independent, then \(\Entr(W) = \Entr(W \mid V)\). Analogously to the unconditional case, we can get an upper bound for the conditional entropy by approximating $\Pr(W \mid V)$ with another distribution $Q$.
\begin{equation}\label{eq:empirical_conditional}
  \Entr_Q(W \mid V) \approx \frac{1}{N} \sum_{i=1}^N \log \frac{1}{Q\left(\tilde{\mathbf{w}}^{(i)} \mid \tilde{\mathbf{v}}^{(i)}\right)}
\end{equation}
where $(\tilde{\mathbf{w}}^{(i)}, \tilde{\mathbf{v}}^{(i)}) \sim \Pr(W, V)$.

\subsection{Systematicity as mutual information} \label{sec:systematicity_as_mi}

Mutual information (\(\MI\)) measures the amount of information (bits) that the knowledge of either form or meaning provides about the other. 
 It is the difference between the entropy and conditional entropy:
\begin{subequations}
  \begin{align}
    \MI(W; V) &\equiv \Entr(W) - \Entr(W \mid V) \label{eq:subeq1} \\
      &\approx \Entr_Q(W) - \Entr_Q(W \mid V)\text{.} \label{eq:subeq2}
  \end{align}
\end{subequations}
Systematicity will thus be framed as (statistically significant) nonzero mutual information \(\MI(V; W)\).


\subsection{Learning $Q$} \label{sec:choosing_q}

Our method relies on decomposing mutual information into a difference of entropies, as shown in \autoref{eq:subeq2}.
We use upper bounds on both the entropy and conditional entropy measures, so our calculated mutual information is an estimate. 

This estimate is as good as our bounds are tight, being perfect when $\Pr(W)$ = $Q(W)$ and  $\Pr(W | V)$ = $Q(W |V)$. 
Still, as we subtract two upper bounds, we cannot guarantee that our MI estimate approaches the real MI from above or below because we do not know which of the entropies' bounds are tighter. There is nothing \emph{principled} that we can say about the result, except that it is consistent.

The procedure for learning the distribution $Q$ is, thus, essential to our method.
We must first define a family of distributions $\Psi$ from which \(Q\) is learned.
\brian[disable]{From Sproat: Suggest using a different symbol than $\mathcal{Q}$. From Brian: not sure why, though perhaps for readability.  maybe a Greek letter?} Then, we learn $Q \in \Psi$
by minimizing the right-hand-size of \autoref{eq:cross-entropy}---which corresponds to maximum likelihood estimation
\begin{equation}
 Q = \arg\inf_{q \in \Psi} \frac{1}{N} \sum_{i=1}^N \log \frac{1}{q\left(\tilde{\mathbf{w}}^{(i)}\right)}\text{.}
\end{equation}
In this work, we employ a state-of-the-art phone-level LSTM language model as our \(\Psi\) to approximate \(\Pr(W)\) as closely as possible.

\subsection{Recurrent neural LM}\label{sec:models}

A phone-level language model (LM) provides a probability distribution over $\Sigma^*$:
\begin{equation}
  \Pr(\mathbf{w}) = \prod\limits_{i=1}^{|\mathbf{w}| + 1} \Pr \left(w_i \mid \mathbf{w}_{<i} \right)\text{.}
\end{equation}

Recurrent neural networks are great representation extractors, being able to model long dependencies---up to a few hundred tokens \cite{khandelwal2018sharp}---and complex distributions $\Pr(w_i \mid \mathbf{w}_{<i})$ \cite{mikolov2010recurrent,sundermeyer2012lstm}. We choose LSTM language models in particular, the state-of-the-art for character-level language modeling \cite{merity2018analysis}.\footnote{ Our tokens are phones rather than graphemes.} 

Our architecture embeds a word---a sequence of tokens $w_i \in \Sigma$---using an embedding lookup table, resulting in vectors $\mathbf{z}_i \in \mathbb{R}^d$. These are fed into an LSTM, which produces high-dimensional representations of
the sequence (hidden states):
\begin{equation}
\mathbf{h}_j = \LSTM\left(\mathbf{h}_{j-1}, \mathbf{z}_j\right), \quad j \in \{1, \ldots, n + 1\}\text{,}
\end{equation}
where \(\mathbf{h}_0\) is the zero vector.
Each hidden state is linearly transformed and fed into a softmax function, producing a 
distribution over the next phone:
\(
  \Pr\left(w_{i} \mid  \mathbf{w}_{< i}\right) = \softmax \left(\textbf{W} \textbf{h}_i + \textbf{b}\right)
\).

\section{Experimental Design}

\subsection{Datasets}


We first analyze the CELEX database \citep{baayen1993celex}, which provides many word types for Dutch, English, and German. In measuring systematicity, we control for morphological variation by only considering monomorphemic words, as in \citet{dautriche2017wordform}. 
Our type-level resource contains lemmata, eliminating the noisy effect of morphologically inflected forms. 
CELEX contains 6040 English, 3864 German, and 3603 Dutch lemmata for which we have embeddings.

While CELEX is a large, well annotated corpus, it only spans three lexically related languages. 
The NorthEuraLex database \citep{dellert2017northeuralex} is thus appealing. It is a lexicon of 1016 \say{basic} concepts, written in a unified IPA scheme and aligned across 107 languages that span 21 language families (including isolates).
\footnote{ We omit Mandarin; the absence of tone annotations leaves its phonotactics greatly underspecified. All reported results are for the remaining 106 languages.} 
 While we cannot restrict NorthEuraLex to monomorphemic words (because it was not annotated by linguists and segmentation models are weak for its low-resource languages), it mainly contains word types for basic concepts---e.g., animal names or verbs---so we are comfortable in the modeling assumption that the words are not decomposable into multiple morphemes.

Unlike \citet{dautriche2017wordform}, who draw lexicons from Wikipedia, or \citet{otis2008phonaesthemes}, we directly use a phone string representation, rather than their proxy of using each language's orthography. This makes our work the first to quantify the interface between phones and meaning in a massively multilingual setting\damian[disable]{er not really - my own 2016 paper covers sound-meaning associations in roughly 4500 languages. Remove?}\tiago[disable]{What about now Damian? I changed `consider' $\rightarrow$ `quantify'}.

\citet{blasi2016sound} is the only large-scale exploration of phonetic representations that we find. They examine 40 aligned concepts over 4000 languages and identify that sound correspondences exist across the vast majority. Their resource \citep{asjp} does not have enough examples to train our language models, and we add to their findings by measuring a relationship \emph{between form and meaning}, rather than form given meaning.

\subsection{Embeddings}

We use pre-trained \textsc{word2vec} representations as meaning vectors for the basic concepts. For CELEX, specific representations were pre-trained for each of the three languages.\footnote{ We use Google's \textsc{word2vec} representations pre-trained in Google News corpus for English, while \textsc{word2vec} was trained using Wikipedia dumps for German and Dutch with default hyper-parameters.}
For NorthEuraLex, as its words are concept aligned, we use the same English vectors for all languages. Pragmatically, we choose English because its vectors have the largest coverage of the lexicon. This does not mean that we assume that semantic spaces across languages to be strictly comparable.  In fact, we would expect that more direct methods of estimating these vectors would be preferable if they were practical.

Note that the methods described above are likely underestimating the semantic systematicity in the data, for a couple of reasons.  First, \textsc{word2vec} and other related methods have been shown to do a better job at capturing general relatedness rather than semantic similarity per se \cite{hill2015simlex}.  Second, our use of the English vectors across the concept-aligned corpora is a somewhat coarse expedient.  To the extent that the English serves as a poor model for the other languages, we should expect smaller MI estimates.  In short, we have chosen easy-to-replicate methods based on commonly used models, rather than extensively tuning our approach for these experiments, possibly at the expense of the size of the effect we observe.


To reduce spurious fitting to noise in the dataset, we reduce the dimensionality of these vectors from the original \(300\) to \(d\) while capturing maximal variance, using principal components analysis (PCA). 

These resulting \(d\)-dimensional vectors are kept fixed while training the conditional language model. Each \(d\)-dimensional vector \(\mathbf{v}\) is linearly transformed to serve as the initial hidden state of the conditional LSTM language model:
\begin{align*}
\mathbf{h}_{0} =&  \mathbf{W}^{(v)}\mathbf{v} + \mathbf{b}^{(v)} \\
\mathbf{h}_j =& \LSTM\left(\mathbf{h}_{j-1}, \mathbf{z}_j\right), \quad j \in \{1, \ldots, n + 1\}\text{.}
\end{align*}
%
We reject morphologically informed embeddings \citep[e.g.,][]{bojanowski2017enriching} because this would be circular: We cannot question the arbitrariness of the form--meaning interface if the meaning representations are constructed with explicit information from the form. This is the same reason that we do not fine-tune the embeddings---our goal is to enforce as clean a separation as possible of model and form, then suss out what is inextricable.

\subsection{Controlling for grammatical category}
The value of \textsc{word2vec} comes from distilling more than just meaning. It also encodes the grammatical classes of words. Unfortunately, this is a trivial source of systematicity: if a language's lemmata for some class follow a regular pattern (such as the verbal infinitive endings in Romance languages), our model will have uncovered something meaningless. Prior work---e.g., \citep{dautriche2017wordform,gutierrez2016finding}---does not account for this\damian[disable]{for instance? I'd ditch this unless explicit refs added}\tiago[disable]{Done!}. To isolate factors like these, we can estimate the mutual information between word form and meaning, while conditioning on a third factor. The expression is similar to \autoref{eq:subeq1}:
\begin{equation}
    \MI(W; V \mid C) \equiv \Entr(W \mid C) - \Entr(W \mid V, C)\text{,}
\end{equation}
where \(C\) is our third factor---in this case, grammatical class.\footnote{ If markers of subclasses within a given part of speech are frequent, these may also emerge.} 

Both CELEX and NorthEuraLex are annotated with grammatical classes for each word. We create a lookup embedding for each class in a language, then use the resulting representation as an initial hidden state to the LSTM ($\mathbf{h}_0 = \mathbf{c}$).
When conditioning on both meaning and class, we concatenate half-sized representations of the meaning (pre-trained) and class to create the first hidden state ($\mathbf{h}_0 = [\mathbf{c}'; \mathbf{W}^{(v)}\mathbf{v}' + \mathbf{b}^{(v)}]$)\damian[disable]{sorry, I don't follow -why this?}\tiago[disable]{We have only `one' initial hidden state for the LSTM, so we have to share it between $V$ and $C$ to condition the LSTM. Is it clearer now with the math notation?}.\looseness=-1


%

\begin{table*}[t]
    \centering
    \begin{tabular}{l c c c c c c c}
    \toprule
        & & \multicolumn{3}{c}{Systematicity} & \multicolumn{3}{c}{Systematicity controlling for POS tags} \\
        \cmidrule(lr{.5em}){3-5}
        \cmidrule(lr{.5em}){6-8}
        Language 
        & $\Entr(W)$
        & $\MI(W ; V)$ & $\Unc(W \mid V)$ & Cohen's $d$ & $\MI(W ; V \mid \text{POS})$  & $\Unc(W \mid V; \text{POS})$ & Cohen's $d$ \\
    \midrule
English & 3.401 & 0.110 & 3.24\% & 0.175 & 0.084  & 2.50\% & 0.133 \\ 
German & 3.195 & 0.168 & 5.26\% & 0.221 & 0.154  & 4.84\% & 0.203 \\ 
Dutch & 3.245 & 0.156 & 4.82\% & 0.222 & 0.089  & 2.84\% & 0.123 \\ 
    \bottomrule
    \end{tabular}
    \caption{Mutual information (in bits per phone), uncertainty coefficients, and Cohen's effect size results for CELEX. Per-phone word--form entropy added for comparison. All mutual information values are statistically significant ($p<10^{-5}$), as tested with a permutation test with $10^5$ permutations.}
    \label{tab:celex}
\end{table*}

\subsection{Hypothesis testing}
We follow \citet{gutierrez2016finding} and \citet{liu2018discovering} in using a permutation test to assess our statistical significance. In it, we randomly swap the sign of $\MI$ values for each word, showing mutual information is significantly positive. Our null hypothesis, then, is that this value should be 0. Recomputing the average mutual information over many shufflings gives rise to an empirical \(p\)-value: asymptotically, it will be twice the fraction of permutations with a higher mutual information than the true lexicon. In our case, we used 100,000 random permutations.


\subsection{Hyperparameters and optimization} \label{sec:optimization}

We split both datasets into ten folds, using one fold for validation, another for testing, and the rest for training.
We optimize all hyper-parameters with 50 rounds of Bayesian optimization---this includes the number of layers in the LSTM, its hidden size, the PCA size $d$ used to compress the meaning vectors, and a dropout probability. Such an optimization is important to get tighter bounds for the entropies, as discussed in \autoref{sec:choosing_q}. We use a Gaussian process prior and maximize the expected improvement on the validation set, as in \citet{snoek2012practical}.
\footnote{ Our implementation is available at \url{https://github.com/tpimentelms/meaning2form}.}

\section{Results and Analysis}

\begin{table*}
    \centering
    \begin{adjustbox}{max width=\textwidth}
    \begin{tabular}{l l r l r}
    \toprule
        Language & Phonestheme & Count & Examples & \(p\)-value \\
    \midrule
Dutch & \prefix{sx} & 110 & schelp, schild, schot, shacht, schaar 
& $<$0.00001 \\ 
& \suffix{@l} & 124 & kegel, nevel, beitel, vleugel, zetel 
& $<$0.00001 \\ 
& \suffix{xt} & 42 & beicht, nacht, vocht, plicht, licht 
& $<$0.00001 \\ 
& \suffix{Op} & 21 & stop, shop, drop, top, bob 
& 0.00068 \\ 
    \midrule
English & \prefix{In} & 33 & infidel, intellect, institute, enigma, interim 
& $<$0.00001 \\ 
& \prefix{sl} & 59 & slop, slough, sluice, slim, slush 
& $<$0.00001 \\ 
& \suffix{kt} & 36 & aspect, object, fact, viaduct, tact 
& 0.00001 \\ 
& \suffix{m@} & 32 & panorama, asthma, trachoma, eczema, magma 
& 0.00002 \\ 
& \suffix{mp} & 44 & stump, cramp, pump, clamp, lump 
& 0.00003 \\ 
& \suffix{@m} & 62 & millennium, amalgam, paroxysm, pogrom, jetsam 
& 0.00007 \\ 
& \prefix{fl} & 64 & flaw, flake, fluff, flail, flash 
& 0.00009 \\ 
& \prefix{bV} & 35 & bum, bunch, bunk, butt, buck 
& 0.00013 \\ 
& \suffix{Qp} & 23 & hop, strop, plop, pop, flop 
& 0.00032 \\ 
& \prefix{gl} & 28 & gleam, gloom, glaze, glee, glum 
& 0.00046 \\ 
& \prefix{sn} & 38 & sneak, snide, snaffle, snout, snook 
& 0.00077 \\ 
& \suffix{n@} & 34 & henna, savanna, fauna, alumna, angina 
& 0.00102 \\ 
& \suffix{\ae{}g} & 23 & swag, shag, bag, mag, gag 
& 0.00107 \\ 
& \prefix{sw} & 43 & swamp, swoon, swish, swoop, swig 
& 0.00112 \\ 
& \prefix{sI} & 78 & silica, secede, silicone, secrete, cereal 
& 0.00198 \\ 
& \suffix{k@} & 22 & japonica, yucca, mica, hookah, circa 
& 0.00217 \\ 
& \prefix{sE} & 34 & shell, sheriff, shelf, chevron, shed 
& 0.00217 \\ 
& \prefix{k@n} & 31 & conceal, condemn, concert, construe, continue 
& 0.00429 \\ 
    \midrule
German & \prefix{g@} & 69 & geschehen, Gebiet, gering, Geruecht, gesinnt 
& $<$0.00001 \\ 
& \suffix{@ln} & 58 & rascheln, rumpeln, tummeln, torkeln, mogeln 
& $<$0.00001 \\ 
& \suffix{ln} & 58 & rascheln, rumpeln, tummeln, torkeln, mogeln  
& $<$0.00001 \\ 
& \suffix{@n} & 801 & goennen, saeen, besuchen, giessen, streiten 
& $<$0.00001 \\ 
& \prefix{In} & 34 & Indiz, indes, intern, innehaben, innerhalb 
& $<$0.00001 \\ 
& \prefix{b@} & 32 & bestaetigen, beweisen, bewerkstelligen, betrachten, beschwichtigen 
& $<$0.00001 \\ 
& \suffix{p@} & 36 & Lampe, Klappe, Kappe, Raupe, Wespe 
& 0.00002 \\ 
& \suffix{S@n} & 24 & dreschen, wischen, mischen, rauschen, lutschen 
& 0.00002 \\ 
& \prefix{Sl} & 39 & schlagen, schlingen, schleifen, schleudern, schluepfen 
& 0.00015 \\ 
& \suffix{k@n} & 76 & backen, strecken, spucken, druecken, schmecken 
& 0.00016 \\ 
& \suffix{ts@n} & 47 & blitzen, schwatzen, duzen, stanzen, einschmelzen 
& 0.00026 \\ 
& \suffix{l@n} & 41 & quellen, prellen, johlen, bruellen, eilen 
& 0.00029 \\ 
& \prefix{ain} & 25 & einstehen, eintreiben, einmuenden, einfinden, eingedenk 
& 0.00033 \\ 
& \suffix{Ix} & 59 & reich, weich, bleich, gleich, Laich 
& 0.00033 \\ 
& \prefix{Sn} & 22 & schnitzen, schnalzen, schnappen, schnurren, schneiden 
& 0.00036 \\ 
& \prefix{Sm} & 23 & schmieren, schmieden, schmunzeln, schmoren, schmeissen 
& 0.00077 \\ 
& \prefix{Sv} & 38 & schweben, schweifen, schwirren, schwellen, schwimmen 
& 0.00124 \\ 
& \suffix{r@n} & 62 & servieren, wehren, sparen, kapieren, hantieren 
& 0.00247 \\ 
& \prefix{br} & 35 & brausen, bremsen, brechen, brennen, brauen 
& 0.00258 \\ 
& \suffix{t@} & 86 & Paste, Quote, Kette, vierte, Sorte 
& 0.00281 \\ 
& \suffix{n@} & 66 & Traene, Tonne, Laterne, Fahne, Spinne 
& 0.00354 \\ 
& \suffix{@rn} & 70 & schillern, schimmern, kapern, knattern, rattern 
& 0.00365 \\  
    \bottomrule
    \end{tabular}
    \end{adjustbox}
    \caption{Discovered phonesthemes, represented as IPA, in Dutch, English, and German, sorted \(p\)-values according to the Benjamini--Hochberg \shortcite{benjamini1995controlling} correction. Count refers to the number of types in our corpus with that affix.}
    \label{tab:german1}
\end{table*}

\subsection{Identifying systematicity} \label{sec:identifying-systematicity}
%



We find statistically significant nonzero mutual information in all three CELEX languages (Dutch, English, and German), using a permutation test to establish significance. This gives us grounds to reject the null hypothesis.
We also find a statistically significant mutual information when conditioning entropies in words' grammar classes. These results are summarized in \autoref{tab:celex}. 

But how much \emph{could} the mutual information have been? A raw number of bits is not easily interpretable, so we provide another information-theoretic quantity, the \term{uncertainty coefficient}, expressing the fraction of bits we can predict given the meaning:
\(
\Unc(W \mid V) = \frac{\MI(W; V)}{\Entr(W)} \text{.}
\)
 The mutual information \(\MI(W; V)\) is upper-bounded by the language's entropy \(\Entr(W)\), so the uncertainty coefficient is between zero and one.\footnote{ Because of our estimation, it may be less than zero.} For the CELEX data, we give the uncertainty coefficients with and without conditioning on part of speech in \autoref{tab:celex}.

By comparing results with and without conditioning on grammatical category, we see the importance of controlling for known factors of systematicity. As expected, all systematicity (mutual information) results are smaller when we condition on part of speech. 
After conditioning, systematicity remains present, though. In English, we can guess about 3.25\% of the bits encoding the phone sequence, given the meaning. In Dutch and German, these quantities are higher.
The effect size of systematicity in these languages, though, is small.



%

%
\subsection{Broadly multilingual analysis} \label{sec:northeuralex}

On the larger set of languages in NorthEuraLex, we see that 87 of the 106 languages have statistically significant systematicity ($p<0.05$), after Benjamini--Hochberg \shortcite{benjamini1995controlling} corrections. When we control for grammatical classes ($\MI(W ; V \mid \textrm{POS})$), we still get significant systematicity across languages ($p<10^{-3}$). A per-language analysis, though, only finds statistical significance for 17 of them, after Benjamini--Hochberg \shortcite{benjamini1995controlling} corrections. This evinces the importance of conditioning on grammatical category; without doing so, we would find a spurious result due to crafted, morphological systematicity. We present kernel density estimates for these results in \Cref{fig:northeuralex_merged} and give full results in \autoref{appendix:northeuralex}. Across all languages, the average uncertainty coefficient was 1.37\% (Cohen's $d$ $0.1936$). When controlling for grammatical classes, though, it was only 0.2\% (Cohen's $d$ $0.0287$).

\begin{figure}
  \begin{center}
    \includegraphics[width=\columnwidth]{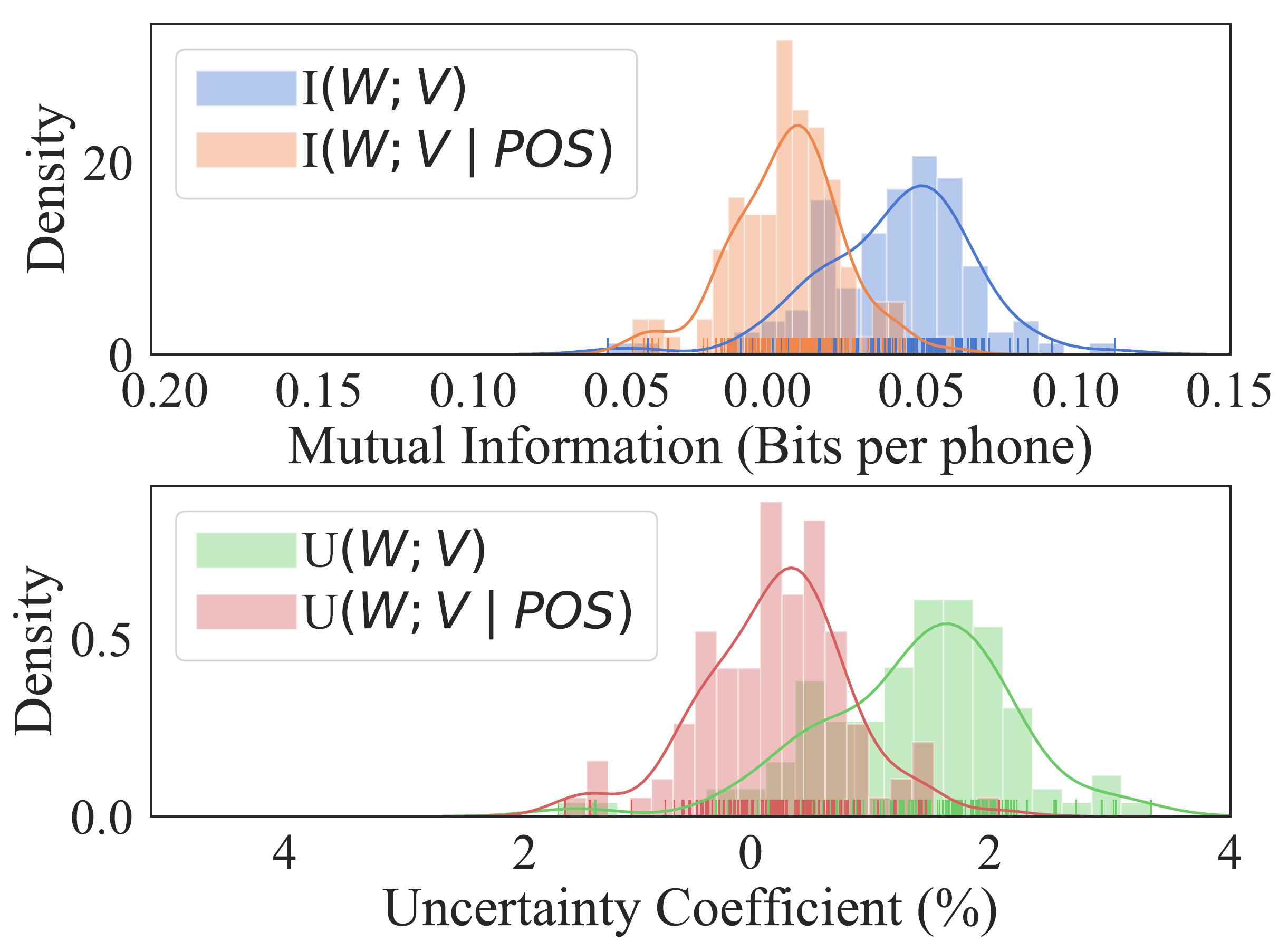}
  \end{center}
  \caption{Mutual information and uncertainty coefficients for each language of NorthEuraLex.}\label{fig:northeuralex_merged}
\end{figure}

%
%

There were only 970 concepts with corresponding \textsc{word2vec} representations in this dataset, and our language models easily overfit when conditioned on these. As we optimize the used number of PCA components ($d$) for these word embeddings, we can check its `optimum' size. The average $d$ across NorthEuraLex languages was only $\approx 22$, while on CELEX it was $\approx 153$. This might imply that the model couldn't find systematicity in some languages due to the dataset's small size---models were too prone to overfitting. 
\tiago{Hey @Damian, could you contribute here with the other reason you gave me of why you think $k$ could be small in NorthEuraLex. The idea that concepts are further apart, so a smaller number of dimensions could be enough to distill semantic difference.}


\subsection{Fantastic phonesthemes and where to find them}%
\label{sec:finding-phonesthemes}

As a phonestheme is, by definition, a sequence of phones that suggest a particular meaning, we expect them to have higher mutual information values when compared to other \(k\)-grams in the lexicon---measured in bits per phone.
To identify that a prefix of length \(k\), \(\mathbf{w}_{\leq k}\), is a phonestheme, we compare it to all such prefixes, being interested in the mutual information \(\MI(W_{\leq k}, V)\). For each prefix in our dataset, we compute the average mutual information over all \(n\) words it appears in.
We then sample $10^5$ other sets of $n$ words and get their average mutual information. Each prefix is identified as a phonestheme with a \(p\)-value of $\frac{r}{10^5}$, where $r$ is how many comparison where it has a lower systematicity than the random sets.\footnote{ While this explanation  is specific to prefixes, we straightforwardly applied this to suffixes by reversing the word forms---e.g. `\textit{banana}' $\mapsto$ `\textit{ananab}'.} \autoref{tab:german1} shows identified phonesthemes for English, Dutch, and German. 
\tiago{@Brian Idea: to control for allogeny here (see Brian's comment), we could compare not to other random sets of words, but random sets of words which all start with the same consonant.}

Inspecting the German  data, it is clear that some of these prefixes and affixes that we find are fossilized pieces of derivational etymology. Further, many of the endings in German are simply the verb ending \suffix{@n} with an additional preceding phone. Dutch and English are less patterned. While we find few examples in Dutch, all are extremely significant. 
It can be argued that two examples (\suffix{@l} and \suffix{xt}) are not semantic markers but rather categorizing heads in the framework of distributed morphology \citep{marantz1993distributed}---suggestions that the words are nouns. Further, in English, we find other examples of fossilized morphology, \mbox{(\prefix{k@n})} and \mbox{(\prefix{In})}. In this sense, our found phonesthemes are related to another class of restricted-application subword: bound morphemes \citep{bloomfield1933language, aronoff1976word, spencer1991morphological}, which carry known meaning and cannot occur alone.

From the list of English prefix phonesthemes we present here, all but \prefix{In} and \prefix{k@n} find support in the literature \citep{hutchins1998psychological,otis2008phonaesthemes,gutierrez2016finding,liu2018discovering}. Furthermore, an interesting case is the suffix \suffix{mp}, which is identified with a high confidence. This might be picking up on phonesthemes \suffix{ump} and \suffix{amp} from \newcite{hutchins1998psychological}'s list.



%
\subsection{Correlation with human judgments}
As a final, albeit weak, validation of our model, we consider how well our computed systematicity compares to human judgments \citep{hutchins1998psychological, gutierrez2016finding, liu2018discovering}. We turn to the survey data of \citet{liu2018discovering}, in which workers on Amazon Mechanical Turk gave a 1-to-5 judgment of how well a word's form suited its meaning. For each of their model's top 15 predicted phonesthemes and 15 random non-predicted phonesthemes, the authors chose five words containing the prefix for workers to evaluate.\footnote{ Of the 150 judgements in their dataset, only 35 were in ours as well, so we restrict our analysis to them. This is a weak signal for our model's validity.} Comparing these judgments to our model-computed estimates of mutual information \(\MI(W_{<2}; V)\), we find a weak, positive Spearman's rank correlation (\(\rho = 0.352\) with \(p = 0.03\)). This  shows that prefixes for which we find higher systematicity---according to mutual information---also tend to have higher human-judged systematicity.

\section{Conclusion}
We have revisited the linguistic question of the arbitrariness---and the systematicity---of the sign. We have framed the question on information-theoretic grounds, 
estimating entropies
by state-of-the-art neural language modeling. We find evidence in 87 of 106 languages for a significant systematic pattern between form and meaning, reducing approximately 5\% of the phone-sequence uncertainty of German lexicons and 2.5\% in English and Dutch, when controlling for part of speech. 

We have identified meaningful phonesthemes according to our operationalization, and we have good precision---all but two of our English phonesthemes are attested in prior work. An avenue for future work is connecting our discovered phonesthemes to putative meanings, as done by \citet{abramova2013automatic} and \citet{abramova2016questioning}.

The low uncertainty reduction suggests that the lexicon is still largely arbitrary. According to the information-theoretic perspective of \citet{monaghan2011arbitrariness}, an optimal lexicon has an arbitrary mapping between form and meaning. If this is true, then a large amount of these benefits do accrue to language; that is, given the small degree of systematicity, we lose little of the benefit.\looseness=-1

\section*{Acknowledgments}
The authors would like to thank Mark Dingemanse, Adina Williams, and the anonymous reviewers for valuable insights and useful suggestions.

\bibliography{naaclhlt2019}
\bibliographystyle{acl_natbib}

\clearpage
\appendix

\section{NorthEuraLex Results}\label{appendix:northeuralex}

\tablehead{%
    \toprule
        Language & $\Entr(W)$ & $\Unc(W \mid V)$ & $\Unc(W \mid V ; \text{POS})$ \\
    \midrule
}
\tabletail{\bottomrule}
\bottomcaption{NorthEuraLex languages and \(p\)-values of systematicity. Bold entries are statistically significant at \(p < 0.05\), after Benjamini--Hochberg \shortcite{benjamini1995controlling} correction.}
\label{tab:northeuralex_mi}
\begin{supertabular}{l r r r}
abk & 2.8432 & \textbf{1.76\%} & -0.26\% \\ 
ady & 3.2988 & \textbf{2.00\%} & 0.50\% \\ 
ain & 3.0135 & 0.54\% & -0.50\% \\ 
ale & 2.5990 & \textbf{1.38\%} & 0.47\% \\ 
arb & 3.0872 & \textbf{1.74\%} & -0.07\% \\ 
ava & 2.8161 & \textbf{2.55\%} & -0.22\% \\ 
azj & 3.0713 & \textbf{1.68\%} & \textbf{1.42\%} \\ 
bak & 3.0652 & \textbf{2.17\%} & 0.44\% \\ 
bel & 3.1212 & \textbf{1.48\%} & -0.37\% \\ 
ben & 3.2638 & \textbf{1.69\%} & \textbf{0.65\%} \\ 
bre & 3.1430 & \textbf{0.57\%} & \textbf{1.43\%} \\ 
bsk & 3.4114 & 0.17\% & 0.10\% \\ 
bua & 2.8739 & \textbf{1.94\%} & 0.02\% \\ 
bul & 3.2150 & \textbf{1.63\%} & 0.19\% \\ 
cat & 3.1536 & \textbf{1.75\%} & 0.11\% \\ 
ces & 3.1182 & \textbf{1.74\%} & 0.19\% \\ 
che & 3.2381 & -1.60\% & 0.62\% \\ 
chv & 3.1185 & 0.43\% & \textbf{0.91\%} \\ 
ckt & 2.8968 & \textbf{1.60\%} & 0.47\% \\ 
cym & 3.2752 & \textbf{1.42\%} & \textbf{0.86\%} \\ 
dan & 3.2458 & \textbf{0.66\%} & 0.57\% \\ 
dar & 3.2124 & \textbf{1.93\%} & -0.37\% \\ 
ddo & 3.2711 & \textbf{2.15\%} & -0.04\% \\ 
deu & 2.9596 & \textbf{1.27\%} & \textbf{0.90\%} \\ 
ekk & 2.9575 & \textbf{0.69\%} & -1.55\% \\ 
ell & 2.9141 & 0.15\% & \textbf{0.89\%} \\ 
enf & 3.0470 & \textbf{3.03\%} & 0.80\% \\ 
eng & 3.2126 & \textbf{0.88\%} & \textbf{0.70\%} \\ 
ess & 2.7369 & \textbf{1.42\%} & 0.29\% \\ 
eus & 3.0070 & \textbf{0.71\%} & -0.57\% \\ 
evn & 2.8434 & \textbf{1.34\%} & 0.64\% \\ 
fin & 2.8996 & \textbf{1.32\%} & 0.23\% \\ 
fra & 3.3423 & \textbf{1.17\%} & -0.32\% \\ 
gld & 2.9055 & \textbf{2.31\%} & 0.26\% \\ 
gle & 3.1450 & 0.51\% & -0.36\% \\ 
heb & 3.1407 & \textbf{1.26\%} & 0.79\% \\ 
hin & 3.0240 & \textbf{1.11\%} & \textbf{0.68\%} \\ 
hrv & 3.0776 & \textbf{2.04\%} & 0.43\% \\ 
hun & 3.2520 & 0.44\% & 0.09\% \\ 
hye & 3.3416 & \textbf{1.84\%} & 0.38\% \\ 
isl & 3.0386 & 0.50\% & -0.71\% \\ 
ita & 2.8409 & \textbf{2.18\%} & 0.57\% \\ 
itl & 3.4332 & \textbf{1.96\%} & 0.27\% \\ 
jpn & 2.8157 & \textbf{1.72\%} & 0.53\% \\ 
kal & 2.5255 & \textbf{1.34\%} & 0.02\% \\ 
kan & 2.8412 & 0.23\% & 0.40\% \\ 
kat & 3.1831 & \textbf{2.04\%} & \textbf{1.06\%} \\ 
kaz & 3.0815 & \textbf{2.19\%} & -0.13\% \\ 
kca & 2.8779 & \textbf{2.93\%} & \textbf{1.40\%} \\ 
ket & 3.3202 & \textbf{0.72\%} & 0.30\% \\ 
khk & 2.9746 & 0.57\% & 0.45\% \\ 
kmr & 3.1292 & \textbf{2.22\%} & 0.26\% \\ 
koi & 3.2419 & 0.57\% & 0.25\% \\ 
kor & 3.1600 & \textbf{1.66\%} & 0.40\% \\ 
kpv & 3.1685 & \textbf{1.71\%} & 0.48\% \\ 
krl & 2.8655 & \textbf{2.19\%} & -0.71\% \\ 
lat & 2.8102 & \textbf{1.36\%} & 0.01\% \\ 
lav & 2.8679 & 0.60\% & -0.10\% \\ 
lbe & 3.0239 & \textbf{0.94\%} & -0.41\% \\ 
lez & 3.3717 & \textbf{3.34\%} & 0.24\% \\ 
lit & 2.8086 & \textbf{1.45\%} & -1.33\% \\ 
liv & 3.0825 & \textbf{1.11\%} & -1.34\% \\ 
mal & 2.6773 & \textbf{1.90\%} & 0.38\% \\ 
mdf & 2.9186 & \textbf{1.24\%} & -0.07\% \\ 
mhr & 2.9952 & \textbf{1.08\%} & \textbf{1.20\%} \\ 
mnc & 2.5750 & \textbf{3.05\%} & -0.03\% \\ 
mns & 2.8001 & \textbf{1.03\%} & 0.18\% \\ 
mrj & 3.1771 & \textbf{1.74\%} & 0.49\% \\ 
myv & 2.8785 & \textbf{1.61\%} & \textbf{0.75\%} \\ 
nio & 2.8985 & \textbf{1.96\%} & \textbf{1.46\%} \\ 
niv & 3.4408 & \textbf{1.46\%} & 0.45\% \\ 
nld & 3.0407 & \textbf{1.56\%} & -0.40\% \\ 
nor & 3.0315 & \textbf{0.68\%} & 0.21\% \\ 
olo & 3.0151 & \textbf{1.38\%} & 0.49\% \\ 
oss & 3.2484 & \textbf{1.42\%} & -0.45\% \\ 
pbu & 3.2840 & \textbf{1.58\%} & -0.05\% \\ 
pes & 2.8443 & \textbf{1.63\%} & -0.17\% \\ 
pol & 3.3167 & \textbf{1.65\%} & 0.27\% \\ 
por & 3.2509 & \textbf{1.19\%} & 0.10\% \\ 
ron & 3.3667 & 0.43\% & -0.99\% \\ 
rus & 3.3538 & \textbf{1.88\%} & 0.17\% \\ 
sah & 3.0002 & -1.29\% & -0.37\% \\ 
sel & 2.8460 & \textbf{1.86\%} & 0.76\% \\ 
sjd & 2.7920 & -0.05\% & 0.30\% \\ 
slk & 3.1928 & \textbf{1.27\%} & 0.46\% \\ 
slv & 2.8685 & \textbf{2.13\%} & -0.40\% \\ 
sma & 2.5011 & \textbf{2.02\%} & -0.14\% \\ 
sme & 2.6746 & \textbf{2.10\%} & -0.17\% \\ 
smj & 2.5975 & \textbf{0.86\%} & -0.52\% \\ 
smn & 2.9281 & \textbf{1.50\%} & 0.22\% \\ 
sms & 2.7608 & \textbf{1.06\%} & -0.56\% \\ 
spa & 2.9777 & \textbf{1.91\%} & \textbf{2.07\%} \\ 
sqi & 3.3473 & 0.22\% & \textbf{0.69\%} \\ 
swe & 2.8600 & \textbf{0.64\%} & -0.44\% \\ 
tam & 2.6851 & -0.19\% & -0.63\% \\ 
tat & 3.1365 & \textbf{1.50\%} & 0.17\% \\ 
tel & 2.8458 & 0.06\% & -1.34\% \\ 
tur & 2.9646 & \textbf{1.93\%} & 0.81\% \\ 
udm & 3.1042 & \textbf{2.72\%} & 0.37\% \\ 
ukr & 3.1135 & \textbf{1.46\%} & 0.48\% \\ 
uzn & 3.0624 & \textbf{1.26\%} & 0.13\% \\ 
vep & 3.2055 & \textbf{2.53\%} & \textbf{1.21\%} \\ 
xal & 3.2090 & \textbf{1.50\%} & 0.51\% \\ 
ykg & 2.9680 & \textbf{1.79\%} & 0.65\% \\ 
yrk & 2.8453 & \textbf{1.97\%} & 0.49\% \\ 
yux & 3.0704 & -0.29\% & -0.18\% \\ 

\end{supertabular}

\vspace{1em}

\tablehead{%
    \toprule
        Language & $\Entr(W)$ & $\Unc(W ; V)$ & $\Unc(W ; V \mid \text{POS})$ \\
    \midrule
}
\tabletail{\bottomrule}
\bottomcaption{NorthEuraLex languages and their uncertainty coefficients. Bold entries are statistically significant at \(p < 0.05\), after Benjamini--Hochberg \shortcite{benjamini1995controlling} correction.}
\label{tab:northeuralex_unc}

\begin{supertabular}{l r r r}
abk & 2.8432 & \textbf{0.0500} & -0.0071 \\ 
ady & 3.2988 & \textbf{0.0661} & 0.0158 \\ 
ain & 3.0135 & 0.0161 & -0.0150 \\ 
ale & 2.5990 & \textbf{0.0358} & 0.0117 \\ 
arb & 3.0872 & \textbf{0.0538} & -0.0020 \\ 
ava & 2.8161 & \textbf{0.0717} & -0.0059 \\ 
azj & 3.0713 & \textbf{0.0517} & \textbf{0.0429} \\ 
bak & 3.0652 & \textbf{0.0666} & 0.0130 \\ 
bel & 3.1212 & \textbf{0.0462} & -0.0110 \\ 
ben & 3.2638 & \textbf{0.0553} & \textbf{0.0206} \\ 
bre & 3.1430 & \textbf{0.0181} & \textbf{0.0444} \\ 
bsk & 3.4114 & 0.0057 & 0.0034 \\ 
bua & 2.8739 & \textbf{0.0558} & 0.0007 \\ 
bul & 3.2150 & \textbf{0.0523} & 0.0060 \\ 
cat & 3.1536 & \textbf{0.0550} & 0.0032 \\ 
ces & 3.1182 & \textbf{0.0543} & 0.0055 \\ 
che & 3.2381 & -0.0519 & 0.0194 \\ 
chv & 3.1185 & 0.0135 & \textbf{0.0282} \\ 
ckt & 2.8968 & \textbf{0.0464} & 0.0131 \\ 
cym & 3.2752 & \textbf{0.0464} & \textbf{0.0275} \\ 
dan & 3.2458 & \textbf{0.0214} & 0.0183 \\ 
dar & 3.2124 & \textbf{0.0621} & -0.0114 \\ 
ddo & 3.2711 & \textbf{0.0702} & -0.0013 \\ 
deu & 2.9596 & \textbf{0.0377} & \textbf{0.0261} \\ 
ekk & 2.9575 & \textbf{0.0203} & -0.0438 \\ 
ell & 2.9141 & 0.0044 & \textbf{0.0252} \\ 
enf & 3.0470 & \textbf{0.0923} & 0.0233 \\ 
eng & 3.2126 & \textbf{0.0284} & \textbf{0.0226} \\ 
ess & 2.7369 & \textbf{0.0388} & 0.0076 \\ 
eus & 3.0070 & \textbf{0.0214} & -0.0166 \\ 
evn & 2.8434 & \textbf{0.0382} & 0.0175 \\ 
fin & 2.8996 & \textbf{0.0384} & 0.0063 \\ 
fra & 3.3423 & \textbf{0.0392} & -0.0104 \\ 
gld & 2.9055 & \textbf{0.0670} & 0.0073 \\ 
gle & 3.1450 & 0.0161 & -0.0111 \\ 
heb & 3.1407 & \textbf{0.0396} & 0.0243 \\ 
hin & 3.0240 & \textbf{0.0336} & \textbf{0.0200} \\ 
hrv & 3.0776 & \textbf{0.0627} & 0.0127 \\ 
hun & 3.2520 & 0.0143 & 0.0029 \\ 
hye & 3.3416 & \textbf{0.0615} & 0.0125 \\ 
isl & 3.0386 & 0.0153 & -0.0208 \\ 
ita & 2.8409 & \textbf{0.0618} & 0.0153 \\ 
itl & 3.4332 & \textbf{0.0674} & 0.0090 \\ 
jpn & 2.8157 & \textbf{0.0485} & 0.0141 \\ 
kal & 2.5255 & \textbf{0.0340} & 0.0005 \\ 
kan & 2.8412 & 0.0066 & 0.0111 \\ 
kat & 3.1831 & \textbf{0.0649} & \textbf{0.0325} \\ 
kaz & 3.0815 & \textbf{0.0676} & -0.0039 \\ 
kca & 2.8779 & \textbf{0.0843} & \textbf{0.0387} \\ 
ket & 3.3202 & \textbf{0.0240} & 0.0100 \\ 
khk & 2.9746 & 0.0170 & 0.0128 \\ 
kmr & 3.1292 & \textbf{0.0694} & 0.0078 \\ 
koi & 3.2419 & 0.0185 & 0.0077 \\ 
kor & 3.1600 & \textbf{0.0524} & 0.0122 \\ 
kpv & 3.1685 & \textbf{0.0542} & 0.0148 \\ 
krl & 2.8655 & \textbf{0.0629} & -0.0195 \\ 
lat & 2.8102 & \textbf{0.0381} & 0.0002 \\ 
lav & 2.8679 & 0.0172 & -0.0027 \\ 
lbe & 3.0239 & \textbf{0.0285} & -0.0119 \\ 
lez & 3.3717 & \textbf{0.1126} & 0.0077 \\ 
lit & 2.8086 & \textbf{0.0409} & -0.0354 \\ 
liv & 3.0825 & \textbf{0.0342} & -0.0401 \\ 
mal & 2.6773 & \textbf{0.0508} & 0.0097 \\ 
mdf & 2.9186 & \textbf{0.0363} & -0.0021 \\ 
mhr & 2.9952 & \textbf{0.0325} & \textbf{0.0348} \\ 
mnc & 2.5750 & \textbf{0.0785} & -0.0006 \\ 
mns & 2.8001 & \textbf{0.0289} & 0.0048 \\ 
mrj & 3.1771 & \textbf{0.0552} & 0.0151 \\ 
myv & 2.8785 & \textbf{0.0463} & \textbf{0.0208} \\ 
nio & 2.8985 & \textbf{0.0569} & \textbf{0.0408} \\ 
niv & 3.4408 & \textbf{0.0504} & 0.0147 \\ 
nld & 3.0407 & \textbf{0.0474} & -0.0118 \\ 
nor & 3.0315 & \textbf{0.0206} & 0.0061 \\ 
olo & 3.0151 & \textbf{0.0415} & 0.0143 \\ 
oss & 3.2484 & \textbf{0.0460} & -0.0140 \\ 
pbu & 3.2840 & \textbf{0.0518} & -0.0017 \\ 
pes & 2.8443 & \textbf{0.0463} & -0.0046 \\ 
pol & 3.3167 & \textbf{0.0547} & 0.0086 \\ 
por & 3.2509 & \textbf{0.0387} & 0.0031 \\ 
ron & 3.3667 & 0.0144 & -0.0322 \\ 
rus & 3.3538 & \textbf{0.0631} & 0.0056 \\ 
sah & 3.0002 & -0.0388 & -0.0111 \\ 
sel & 2.8460 & \textbf{0.0528} & 0.0207 \\ 
sjd & 2.7920 & -0.0013 & 0.0082 \\ 
slk & 3.1928 & \textbf{0.0406} & 0.0139 \\ 
slv & 2.8685 & \textbf{0.0611} & -0.0111 \\ 
sma & 2.5011 & \textbf{0.0505} & -0.0033 \\ 
sme & 2.6746 & \textbf{0.0562} & -0.0043 \\ 
smj & 2.5975 & \textbf{0.0223} & -0.0129 \\ 
smn & 2.9281 & \textbf{0.0439} & 0.0061 \\ 
sms & 2.7608 & \textbf{0.0292} & -0.0149 \\ 
spa & 2.9777 & \textbf{0.0568} & \textbf{0.0599} \\ 
sqi & 3.3473 & 0.0073 & \textbf{0.0226} \\ 
swe & 2.8600 & \textbf{0.0182} & -0.0124 \\ 
tam & 2.6851 & -0.0050 & -0.0167 \\ 
tat & 3.1365 & \textbf{0.0471} & 0.0050 \\ 
tel & 2.8458 & 0.0017 & -0.0374 \\ 
tur & 2.9646 & \textbf{0.0574} & 0.0234 \\ 
udm & 3.1042 & \textbf{0.0843} & 0.0110 \\ 
ukr & 3.1135 & \textbf{0.0456} & 0.0142 \\ 
uzn & 3.0624 & \textbf{0.0386} & 0.0039 \\ 
vep & 3.2055 & \textbf{0.0812} & \textbf{0.0374} \\ 
xal & 3.2090 & \textbf{0.0482} & 0.0156 \\ 
ykg & 2.9680 & \textbf{0.0532} & 0.0186 \\ 
yrk & 2.8453 & \textbf{0.0561} & 0.0133 \\ 
yux & 3.0704 & -0.0088 & -0.0054 \\ 

    \end{supertabular}

\end{document}